\documentclass[letterpaper, 10 pt, conference]{ieeeconf}
\IEEEoverridecommandlockouts
\usepackage{times}
\usepackage{soul}
\usepackage{url}
\usepackage[hidelinks]{hyperref}
\usepackage[utf8]{inputenc}
\usepackage{graphicx}
\usepackage{amsmath}
\usepackage{amssymb}
\usepackage{booktabs}
\usepackage{algorithm}
\usepackage{algorithmic}
\usepackage{multirow}
\usepackage{array}
\usepackage{xcolor}

\title{\LARGE \bf
Task-Motion Planning for Safe and Efficient Urban Driving
}

\author{Yan Ding, Xiaohan Zhang, Xingyue Zhan, Shiqi Zhang
\thanks{The authors are with the Department of Computer Science, SUNY Binghamton. Emails: \{yding25; xzhan244; xzhan215; zhangs\}@binghamton.edu}
}

\begin{document}
\maketitle
\begin{abstract}

Autonomous vehicles need to plan at the task level to compute a sequence of symbolic actions, such as merging left and turning right, to fulfill people's service requests, where efficiency is the main concern. 
At the same time, the vehicles must compute continuous trajectories to perform actions at the motion level, where safety is the most important. 
Task-motion planning in autonomous driving faces the problem of maximizing task-level efficiency while ensuring motion-level safety. 
To this end, we develop algorithm Task-Motion Planning for Urban Driving (TMPUD) that, for the first time, enables the task and motion planners to communicate about the safety level of driving behaviors. 
TMPUD has been evaluated using a realistic urban driving simulation platform. 
Results suggest that TMPUD performs significantly better than competitive baselines from the literature in efficiency, while ensuring the safety of driving behaviors.

\end{abstract}

\section{Introduction}
Autonomous driving technologies have the great potential of reshaping urban mobility in people's daily life~\cite{bonnefon2016social,geiger2012we,maurer2016autonomous}. 
To be deemed useful, autonomous vehicles\footnote{Autonomous vehicles are referred to as vehicles for simplicity in the following sections of this paper.} must be time-efficient in accomplishing service tasks, which frequently requires symbolic actions such as ``\emph{Merge left, go straight, turn left, and park right''}, while at the same time ensuring safety in executing such actions on the road~\cite{haboucha2017user,koopman2017autonomous,cao2018analysing}.

\begin{figure}
\vspace{0.8em}
\centering
\includegraphics[width=1\columnwidth]{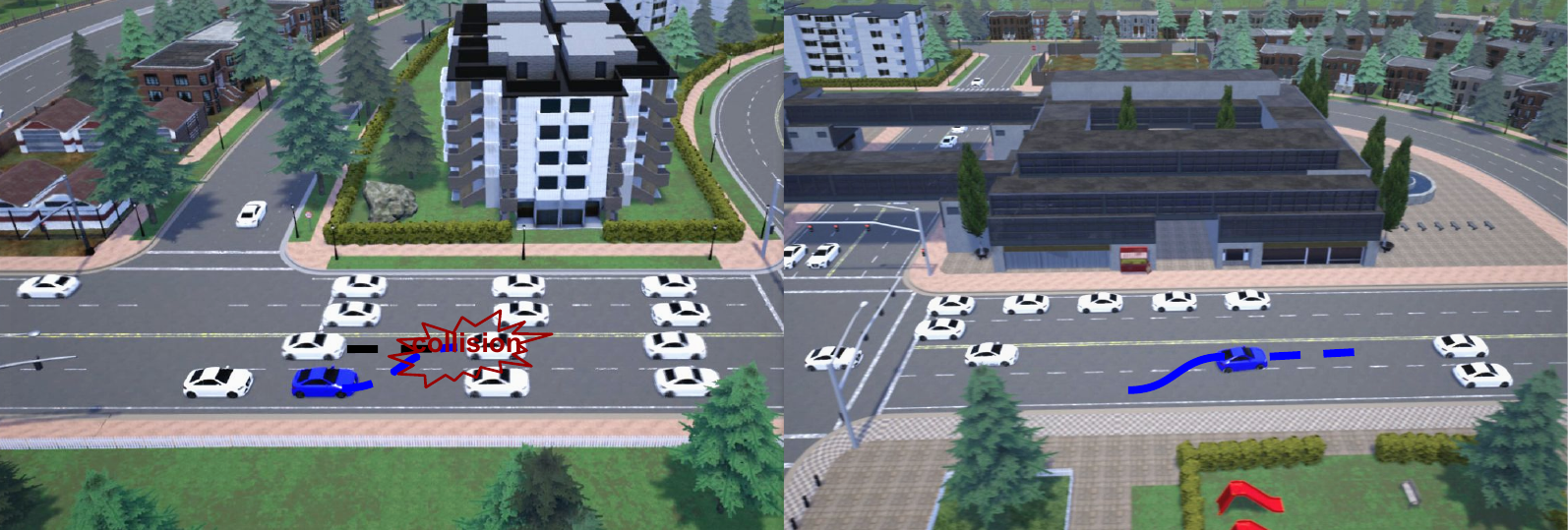}
\caption{\textbf{Left:} a risky situation for the vehicle (blue) to merge left due to the busy traffic. 
\textbf{Right: }
a safe situation for the vehicle to merge left. 
The goal of TMPUD (ours) is to enable the motion level to take symbolic actions from, and communicate safety to the task level toward efficient and safe autonomous driving behaviors. 
} \label{fig:collision_merge}
\end{figure}

Generally, autonomous vehicles need to plan at the \emph{task level} to compute a sequence of symbolic actions toward fulfilling service requests from people. 
In this process, how the actions are implemented in the real world is out of consideration at the task level. 
At the same time, vehicles must plan at the \emph{motion level} to compute continuous trajectories, and desired control signals (e.g., for steering, accelerating, and braking) to implement the symbolic actions. 
While the task planner hopes that all the symbolic actions can be implemented by the vehicles, there is the safety concern that must be considered at the motion level.  
For instance, lane-changing behaviors can be dangerous in heavy traffic. 
Fig.~\ref{fig:collision_merge} shows two situations that are dangerous (\textbf{Left}) and safe (\textbf{Right}) to a vehicle, respectively. 

Although task planning (frequently referred to as behavior planning in autonomous driving~\cite{paden2016survey}) and motion planning have been individually conducted in autonomous driving, there is little research from the literature focusing on the interaction between task and motion levels. 
\emph{There is the critical need of developing algorithms to bridge the gap between task planning and motion planning to help vehicles improve the task-completion efficiency while ensuring the safety of driving behaviors.}

The robotics community has studied the integration of task and motion planning, mostly in manipulation domains~\cite{srivastava2014combined,kim2017learning,garrett2018ffrob,lo2018petlon}.
In comparison to those domains, autonomous driving algorithms must consider the uncertainty from the ego vehicle, and the surrounding objects (including other vehicles) on the road. 
The uncertainty must be quantitatively evaluated at the motion level, and taken into consideration for planning at the task level. 
For instance, \emph{when the left lane is busy and missing the next crossing does not introduce much extra distance, the task planner should avoid forcing the vehicle to merge left.} 
Such behaviors are possible, only if the interactions between task and motion levels are enabled.


In this paper, we develop \textbf{Task-Motion Planning for Urban Driving (TMPUD)} for efficient and safe autonomous urban driving. 
TMPUD, for the first time, enables the interaction between task and motion planners through enabling the motion-level safety estimation and task-level replanning capabilities. 
The \textbf{contribution of this research} is two-fold, including the new safety estimator, and the TMPUD algorithm. 
We have implemented and evaluated TMPUD using CARLA, an autonomous driving platform for simulating urban driving scenarios~\cite{dosovitskiy2017carla}. 
Results suggest TMPUD improves both safety and efficiency, in comparison to two baseline methods from the literature~\cite{chen2015task,srivastava2014combined}. 


\begin{figure*}
\vspace{0.5em}
\centering
\includegraphics[width=1.8\columnwidth]{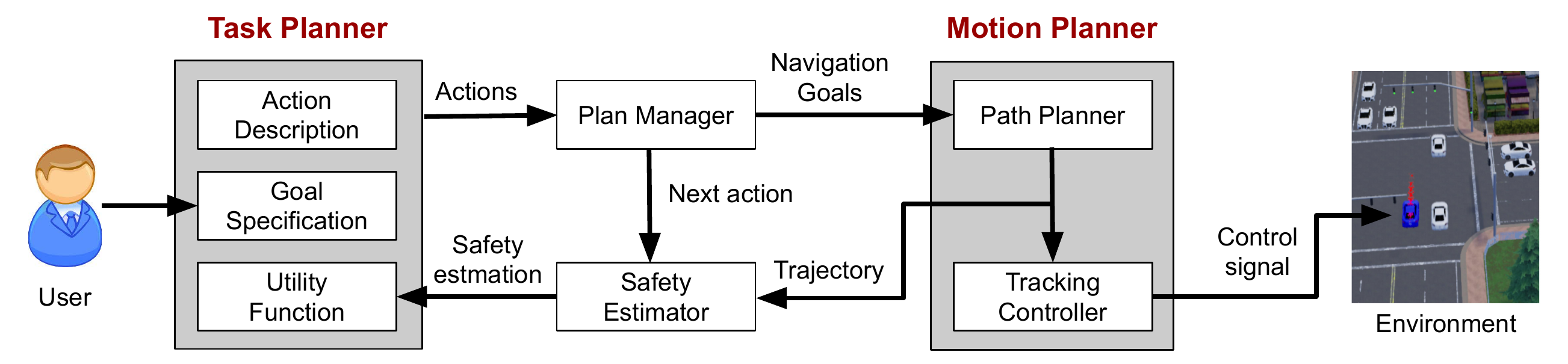}
\caption{An overview of algorithm TMPUD that consists of four components, i.e., \emph{task planner}, \emph{plan manager}, \emph{motion planner} and \emph{safety estimator}. 
\emph{Task planner} includes components of goal specification, action description, and utility function, where users' service requests are received by the \emph{goal specification} component. 
Task planner computes a sequence of symbolic actions that are passed to the \emph{plan manager}. 
The plan manager generate navigation goals (i.e., a pair of two poses) to \emph{path planner}, which is then used for computing a continuous trajectory for connecting 2D poses. 
The trajectory will be used in the two components of \emph{safety estimator} and \emph{tracking controller}. 
\emph{Safety estimator} uses this trajectory to estimate actions' safety levels, and then the utility function in task planner can be updated accordingly. 
\emph{Tracking controller} computes the desired control signals to drive the vehicle to follow the trajectory from the path planner.} 
\label{fig:overview}
\end{figure*}

\section{Related Work}
We summarize three research areas that are the most relevant to this research, namely motion planning in autonomous driving, task planning in autonomous driving, and the integration of task and motion planning. 

\vspace{.8em} 
\noindent \textbf{Motion-Level Planning for Autonomous Driving:} 
Safety is of the most importance at the \emph{motion level}, and highly relies on the motion-level controllers. 
Early research in robotics (mostly on manipulation problems) has developed a ``safe set'' algorithm to avoid unsafe situations in human-robot interactions~\cite{liu2014control}, where it offers a theoretical guarantee of safety. 
That algorithm has been improved to further account for the uncertainty from the real world~\cite{liu2015safe}, where both efficiency and safety were modeled in human-robot interface scenarios. 
Those methods focused on robot manipulation domains, where it is frequently assumed the acting agent being the only one that makes changes in the world, and hence are not applicable to autonomous driving domains. 

More recently, within the autonomous driving context, researchers have developed a series of learning and decision making methods to enable motion planners to learn safe behaviors~\cite{chen2019model, chen2020interpretable, chen2019autonomous, liu2016enabling, bi2019learning, paxton2017combining}. 
The above-mentioned methods (in robotics and autonomous driving) mainly focused on motion-level behaviors, and did not look into how motion-level behaviors can be sequenced at the task level to accomplish complex driving tasks. 
By contrast, TMPUD supports the interaction between task and motion levels, and aims at improving both safety and efficiency of autonomous driving behaviors. 

\vspace{.8em} 
\noindent \textbf{Task-Level Planning for Autonomous Driving:} 
Task planning has been applied to autonomous driving. 
For instance, one of the earliest works on this topic demonstrated that task planning techniques enable vehicles to complete complex tasks, such as to avoid temporary roadblocks~\cite{chen2015task} (we use this approach as a baseline in experiments). 
However, their work did not consider costs of driving behaviors, and hence performs poorly in task-completion efficiency. Similarly, task planners in \cite{lim2009behavior, xiu2010behavior, wei2014behavioral} have \emph{no interaction} with the motion planner, and safety was not modeled in generating the driving behaviors.

Moving forward, more recent research has enabled vehicles to periodically verify the task sequences and motion trajectories against the actual traffic situation~\cite{chen2016combining}. 
In case of possible dangers detected at the motion level, re-planning is triggered at the task level. 
The main limitation of their work is that the triggering is deterministic, and highly depends on a safety threshold. 
The threshold must be set beforehand to ensure safety, which frequently produces over-conservative behaviors, and significantly reduces the task-level efficiency. 


\vspace{.8em} 
\noindent \textbf{Task and Motion Planning:}
Researchers have integrated task and motion planning in robotics, where the primary domain is robot manipulation~\cite{cambon2009hybrid, wolfe2010combined, nau2003shop2, erdem2011combining, srivastava2014combined}. 
Research on manipulation is mostly concerned with the motion-level feasibility, e.g., in grasping and ungrasping behaviors, and accomplishing high-level tasks, such as stacking objects. 
Those methods did not consider the uncertainty from other agents (e.g., vehicles on the road). 
As a result, their systems produce over-optimistic (and hence risky) behaviors, assuming no other agents making changes in the world, and are not applicable to autonomous driving domains. 

A journal paper has surveyed frameworks for autonomous driving~\cite{paden2016survey}, including works that plan at both task and motion levels. 
However, their motion planners do not provide any feedback to the task level, except for infeasible actions. 
In comparison, our TMPUD algorithm supports motion-level safety evaluation, and enables the task planner to dynamically adjust high-level plans to account for current road conditions toward accomplishing long-term driving tasks.

\section{Background}
\label{background}

We very briefly summarize task planning and motion planning, the two building blocks of this research. 

\vspace{.8em}
\noindent \textbf{Task Planning:} 
A task planning domain is specified by $D^t$, including a set of states, $S$, and a set of actions, $A$. We assume a factored state space such that each state $s \in S$ is defined by the values of a fixed set of variables; each action $a \in A$ is defined by its preconditions and effects. A utility function maps the state transition to a real number, which takes both \emph{cost function} $Cost(\langle s,a,s' \rangle)$ and \emph{safety function} $Safe(\langle s,a,s' \rangle)$ into account. 
Specifically, the cost and safety functions respectively reflect the cost and safety of conducting action $a$ in state $s$. 

Given domain $D^t$ and a task planning problem, we want to compute a plan $p \in P$, starting from an initial state $s^{init} \in S$ and finishing in a goal state $s^g \in S$. 
A plan $p$ consists of a sequence of transitions that can be represented as: $p = \langle s_0,a_0, \cdots, s_{N-1},a_{N-1},s_N\rangle$, where $s_0= s^{init}$, $s_N = s^g$ and $P$ denotes a set of satisfactory plans. 
Task planner $P^t$ can produce an optimal plan $p^*$ among all satisfactory plans, where $\gamma$ is a constant coefficient and $\gamma > 0$. 
$$
    \small{p^* = \mathop{\arg\min}\limits_{p\in P}   \!\!\!\!   \sum_{\langle s,a,s'\rangle \in p} \!\!\!\!   [Cost(\langle s,a,s' \rangle)+\frac{\gamma}{1+e^{Safe(\langle s,a,s'\rangle)-1}}]}
$$

\vspace{.8em}
\noindent \textbf{Motion Planning:}  A motion planning domain is specified by $D^m$, where we directly search in 2D space constrained by the urban road network. Some parts of the space are designated as free space, and the rest are designated as obstacles. The 2D space is represented as a region in Cartesian space such that the position and orientation of the vehicle can be uniquely represented as a pose, denoted by $x$. 

Given domain $D^m$, a motion planning problem can be specified by an initial pose $x^{i}$ and a goal pose $x^{g}$. The motion planning problem is solved by a motion planner $P^m$ consisting of path planner and tracking planner into two phases. In the first one, a path planner computes a collision-free trajectory $\xi$ connecting pose $x^{i}$ and pose $x^{g}$ taking into account any motion constraints on the part of the vehicle with minimal trajectory length. In the second one, a tracking controller computes desired control signals to drive the vehicle to follow the computed trajectory. Due to the fundamental difference between representations at task and motion levels, in line with past research~\cite{cambon2009hybrid, erdem2011combining, srivastava2014combined, lo2018petlon}, we use a \emph{state mapping function}, $f: X = f(s)$, to map the symbolic state $s$ into a set of feasible poses $X$ in continuous space, for motion planner to sample from. We assume the availability of at least one pose $x \in X$ in each state s, such that the vehicle is in the free space of $D^m$. If it is not the case, the state $s$ is declared infeasible.

\section{Algorithms}\label{sec:alg}
In this section, we present our main contribution of this research, including two algorithms for safety estimation, and efficient and safe urban driving.

\subsection{Safety Estimation}\label{sec:safety}
Safety estimation aims at computing the safety level, $Safe(\langle s, a, s' \rangle)$, of the motion-level implementation of a symbolic action $\langle s,a,s'\rangle$. 
The goal of computing the safety value is to enable the task planner to incorporate the road condition into the process of sequencing high-level actions toward accomplishing complex driving tasks. 

\vspace{.8em}
\noindent \textbf{Terminology:}  
To perform symbolic action $\langle s,a,s'\rangle$, we use a motion planner to compute a sequence of continuous control signals, i.e., acceleration $\delta \in \Delta$ and steer angle $\theta \in \Theta$, to drive the vehicle following the planned trajectory, while ensuring no collision on the road. 
Sets $\Delta$ and $\Theta$ denote the operation specification of the controller, which generally depends on the adopted motion planner and the ego vehicle itself. 
Let $U_S(t)$ (mathematically $U_S(t) \subset \Delta \times \Theta$) specify a \emph{safe control set} at time $t$, in which all elements, denoted by $u(t)=\langle \delta, \theta \rangle$, are \emph{safe} for an ego vehicle to perform at time $t$. 
Intuitively, the size of safe control set $U_S$ reflects the safety level. 
For instance, when $|U_S|$ is very small, meaning that very few control signals are safe, the vehicle can only be operated in very particular ways, indicating the safety level in general is low. 
Accordingly, \emph{we use the probability of elements sampled from set $\Delta \times \Theta$ being located in the safe set $U_S$ to represent the safety value of action $\langle s, a, s' \rangle$}.

\begin{algorithm}[t]
\caption{Safety Estimation}\label{alg_estimator}
\begin{algorithmic}[1]
\REQUIRE Symbolic action $\langle s,a,s'\rangle$, state mapping function $f$, motion planner $P^m$, control operation sets $\Delta$ and $\Theta$
\STATE Sample initial and goal poses, $x \leftarrow f(s)$ and $x'\leftarrow f(s')$, given action $\langle s,a,s'\rangle$, and $f$. \label{line1:sample}
\STATE Compute a collision-free trajectory, $\xi^E$, using \! $P^m$, where $\xi^E(t_1) \! =\! x,\xi^E(t_2) \! =\! x' \! ,$ and $\![t_1,t_2]\!$ is the horizon \label{line1:collision-free_traj}
\STATE Predict trajectory $\xi^S_i$ for the $i$th surrounding vehicle, where $i\in [1,\cdots,N]$, and $[t_1,t_2]$ is the horizon
\label{line1:predict_traj}
\WHILE{for each vehicle $V_i$}\label{line1:while}
\STATE Compute safe control set $U^S_i(t)$ between the ego vehicle and vehicle $V_i$ at time $t \in [t_1,t_2]$, where $U^S_i(t) \subset \Delta \times \Theta$ and $t = t_1 + \omega \times i, i \leq \lfloor \frac{(t_2-t_1)}{\omega}\rfloor$
\label{line1:compute_safety}
\STATE Sample $M$ elements $\langle \delta,\theta \rangle$ randomly from set $\Delta \times \Theta$ and compute the probability $o_i(t)$ of the elements falling in set $U^S_i(t)$
\label{line1:sample_safety}
\STATE Convert a list of estimated safety values, $\{o_i(t)\}$, into a scalar value $o_i^*$ using Eqn.~\ref{eqn:convert}
\label{line1:convert}
\ENDWHILE\label{line1:endwhile}
\STATE \textbf{return} $min\{o_i^*, i=1,\cdots,N\}$
\label{line1:choose_mini}
\end{algorithmic}
\end{algorithm}

\vspace{.8em}
\noindent \textbf{Safety Estimation Algorithm:}  
Algorithm~\ref{alg_estimator} summarizes the procedure of our safety estimation algorithm. 
The input includes symbolic action $\langle s,a,s' \rangle$, stating mapping function $f$, motion planner $P^m$ consisting of path planner and tracking controller, and the controller's operation specification sets $\Delta$ and $\Theta$. 
The output is the estimated safety value $Safe(\langle s,a,s'\rangle)\in [0.0, 1.0]$. 

Lines~\ref{line1:sample}-\ref{line1:predict_traj} aim to obtain the short-period trajectories of the ego and surrounding vehicles, where $V_i, i\in [1,\cdots N]$, is the $i$th vehicle within the ego vehicle's sensing range. 
More specifically, we first sample a pair of feasible initial and goal poses for the symbolic actions using the state mapping function (Line~\ref{line1:sample}). 
Taking these two poses as input, the motion planner then computes a continuous trajectory for our ego vehicle for a short period of time $[t_1,t_2]$ (Line~\ref{line1:collision-free_traj}), where $t_1$ is the current time, and $t_2=t_1+T$ indicates the time horizon of the ego vehicle. 
We predicate surrounding vehicles' trajectories, assuming their linear and angular speeds being stationary (Line~\ref{line1:predict_traj}), though there are more advanced methods~\cite{houenou2013vehicle,ammoun2009real}, which is beyond the scope of this research. 

Lines~\ref{line1:while}-\ref{line1:endwhile} present a control loop that computes the safety estimation between the ego vehicle and the surrounding vehicles $V_i,~\textnormal{wehre~} i\in[1,\cdots,N]$, given that the ego vehicle is performing action $\langle s,a,s'\rangle$ at the motion level. 
We compute a safe control set $U_i^S(t)$, similar to~\cite{liu2016enabling}, that includes all safe control signals with regard to vehicle $V_i$ at time $t$ (Line~\ref{line1:compute_safety}). 
Parameter $\omega$ controls the sampling interval. 
In Line~\ref{line1:sample_safety}, we randomly sample $M$ elements from the set $\Delta \times \Theta$, and compute probability $o_i(t)$ of the sampled elements falling in set $U^S_i(t)$. 
We convert a list of values of safety estimation $\{o_i(t)\}$ into a single value $o_i^*$ using eqn.\ref{eqn:convert}, where $max$ and $mean$ are two functions to calculate the maximum and mean value of a list, respectively (Line~\ref{line1:convert}).
Although all surrounding vehicles can potentially introduce risks to the ego vehicle, we assume the ego vehicle only considers the most dangerous vehicle. 
Accordingly, Line~\ref{line1:choose_mini} is used for selecting the minimum value, $o_i^*, i \in [1, \cdots, N]$, as the overall safety value: 

\begin{equation}\label{eqn:convert}
    o_i^* = \frac{max_{t \in \mathcal{T}}\{o_i(t)\}+mean_{t \in \mathcal{T}}\{o_i(t)\}}{2}
\end{equation}
where $\mathcal{T}=t_1 + \omega \times i,~~0 \leq i \leq \frac{(t_2-t_1)}{\omega}$


\subsection{TMPUD}
\label{sec:TMPUD}
Our motion planner $P^m$ computes both costs (trajectory lengths) and safety values of the ego vehicle's navigation actions, which have been discussed in Section~\ref{sec:safety}. 
Here, we focus on the main contribution of this work on enabling interactive task-motion planning for urban driving.

\vspace{.8em}
\noindent \textbf{Terminology:}  
We use $s^{init}$ to represent the initial state of the ego vehicle, and 
the goal (\emph{service request} from people) is specified using $s^{g}$. 
Our task planner $P^{t}$ computes a sequence of symbolic actions, and it requires two functions that are initialized and updated within the algorithm, including cost function $Cost$, and safety estimation function $Safe$.
Motion planner $P^m$ is used for computing motion trajectories, and generating control signals to move the ego vehicle. 
The state mapping function $f$ is used for mapping symbolic states to 2D coordinates in continuous spaces.

\begin{algorithm}[t]
\caption{TMPUD algorithm}\label{alg_TMPUD}
\begin{algorithmic}[1]
\REQUIRE Initial state $s^{i}$, goal specification $s^{g}$, task planner $P^t$, state mapping function $f$, motion planner $P^m$, and safety estimator (Algorithm 1)
\STATE Initialize cost function $Cost$ with sampled poses $x \in f(s)$: $Cost(\langle s,a,s'\rangle)\leftarrow$ $A^*(x,x')$
\STATE Initialize safety estimation $Safe(s,a,s') \leftarrow 1.0$ 
\STATE Compute an optimal task plan $p$ using $Cost$ and $Safe$ functions: $p \leftarrow P^t(s^{init}, s^g, Cost, Safe)$, where 
$p = \langle s^{init},a_0,s_1,a_1,\cdots,s^g \rangle$
\WHILE{Plan $p$ is not empty}
\STATE{Extract the first action of $p$, $\langle s,a,s'\rangle$, and compute safety value $\mu$ using \textbf{Algorithm 1}}
\STATE Update $Safe$ function: $Safe(\langle s,a,s'\rangle) \leftarrow \mu$ and $Cost$ function: $Cost(\langle s,a,s'\rangle)\leftarrow$ $A^*(x,x')$
\STATE{Generate a new plan: $p' \leftarrow P^t(s, s^g, Cost, Safe)$ }
\IF{$p'$ $==$ $p$}
\STATE $x' \leftarrow f(s')$
\WHILE{$x~!\!= x'$}
\STATE Call motion planner $\langle \delta, \theta \rangle \leftarrow P^m(x,x')$ 
\STATE Execute the control signal $\langle \delta, \theta \rangle$
\STATE Update the vehicle's current pose $x$
\ENDWHILE
\STATE{Remove the tuple $\langle s, a\rangle$ from plan $p$}
\ELSE
\STATE{Update current plan $p \leftarrow p'$}
\ENDIF
\ENDWHILE
\end{algorithmic}
\end{algorithm}

\vspace{.8em}
\noindent \textbf{The TMPUD Algorithm:}  
Algorithm~\ref{alg_TMPUD} summarizes the procedure of TMPUD. 
It starts by initializing the cost and safety estimation functions (Lines 1 and 2). 
Cost function $Cost$ is initialized using A star algorithm provided by CARLA, as shown in Line 1. 
In Line~2, TMPUD optimistically initializes the safety estimation function by setting $1.0$ to all actions, indicating all task-level actions are completely safe. 
After that, an optimal task plan, $p^* = \langle s^{init},a_0,s_1, \cdots, s^g\rangle$, is computed in Line 3. 
The head and tail elements of the plan, $s^{init}$ and $s^g$, correspond to the initial and goal poses respectively. 

Lines 4-19 form TMPUD's main control loop that enables the interaction between task and motion planners. 
The loop's termination condition is the task-level plan being empty, i.e., the goal has been achieved (Line 4). 
Specifically, TMPUD estimates the safety level, $\mu$, of action $\langle s, a, s'\rangle$ (Line 5). 
Functions $Safe$ and $Cost$ are updated using $\mu$ and $A*$ search in Line 6.
Then a new optimal plan $p'$ is computed in Line 7. 
Lines~8-18 is for plan monitoring and action execution. 
If the task planner suggests the same plan (Line 8), the vehicle will continue to execute action $a$ at the motion level. The goal state is sampled from state mapping function in Line 9. Line 10-14 is a loop to execute the action. Specifically, the motion planner will compute and execute a desired control signal $\langle \delta, \theta \rangle$ repeatedly until the vehicle reaches the goal pose (Line 10). The vehicle's current pose $x$ will be updated after each execution (Line 13). After completing the operation, the tuple $\langle s, a\rangle$ will be removed from the plan $p$ (Line 15). On the contrary, if the task planner suggests a new plan $p'$ different from the plan $p$, the \emph{currently optimal} $p'$ will replace the \emph{non-optimal} plan $p$ (Line 17).

\subsection{Algorithm Instantiation}\label{sec:instantiation}
\noindent\textbf{Task Planner:}
Our task planner $P^t$ is implemented using Answer Set Programming (ASP), which is a popular declarative language for knowledge representation and reasoning, and ASP has been used for task planning~\cite{lifschitz2002answer,amiri2019augmenting,lo2018petlon,jiang2019task}. 
For example, predicate \texttt{leftof(La1,La2)} can be used to specify lane \texttt{La1} being on the left of lane \texttt{La2}. 
We model five driving actions, including \texttt{mergeleft}, \texttt{mergeright}, \texttt{forward}, \texttt{turnleft}, and \texttt{turnright}. 
For instance, action \texttt{mergeright} can be used to help the vehicle merge to the right lane, where constraints, such as ``\texttt{changeright}'' is allowed only if there exists a lane on the right, have been modeled as well. 

\vspace{.8em}
\noindent\textbf{Motion Planner:}
At the motion level, path planner firstly generates a desired continuous trajectory with the minimal traveling distance using $A*$ search. 
The trajectory includes a set of waypoints (each in the form of a pair of $x-y$ coordinate and orientation), and the trajectory is delivered to the tracking controller, along with the vehicle's current pose and speed. 
The controller uses a proportional-integral-derivative (PID) controller~\cite{emirler2014robust} to generate control signals, e.g.,  for steering, throttle, and brake. 
PID controller is very popular due to its simplicity, flexibility, and robustness. 

\section{Experiments}
\label{sec:exp}


We use CARLA, an open-source 3D urban driving simulator~\cite{dosovitskiy2017carla} in this research. 
CARLA has been developed to support development, training, and validation of autonomous driving systems. 
Compared to other simulation platforms, e.g.,~\cite{shah2018airsim,koenig2004design}, CARLA provides open digital assets (urban layouts, buildings, vehicles) that were created for this purpose and can be used freely.



\begin{figure}[t]
\vspace{.5em}
\centering
\includegraphics[width=1\columnwidth]{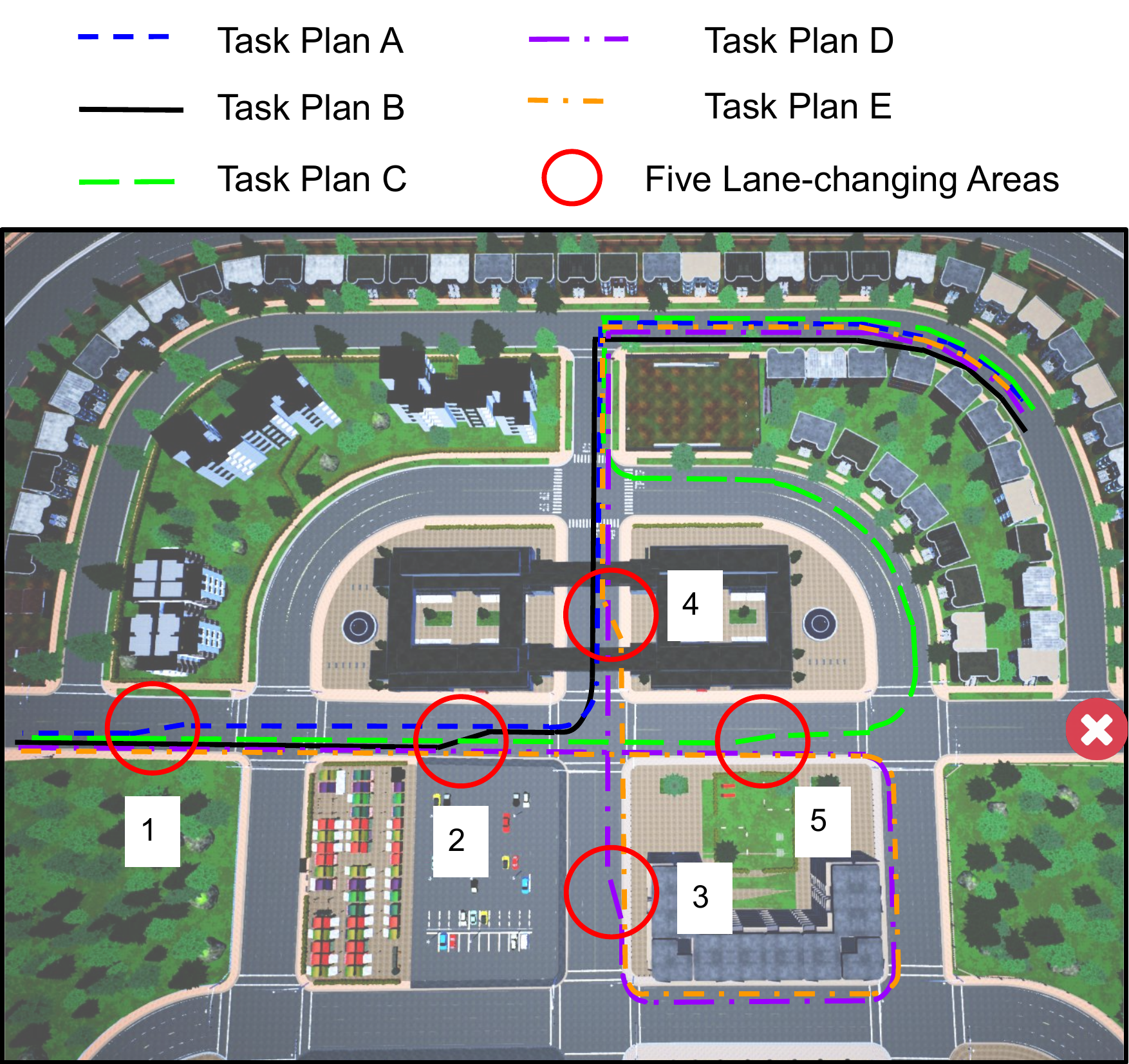}
\caption{An illustrative example, where the vehicle is tasked with driving from the very left to the top-right area. 
The vehicle needs to compute plans at both task and motion levels. 
The vehicle starts with executing \emph{Plan A} (blue color). 
While it is getting close to \emph{Area 1} (a red circle), the motion-level safety estimator reports a low safety value based on the local road condition. 
This computed safety value is incorporated into task planner's cost function. 
Using the updated cost function, the task planner re-computes an optimal plan (\emph{Plan B}), and suggests the vehicle to \emph{go straight} and \emph{merge left in \emph{Area 2}}.
The interaction between task and motion levels, supported by TMPUD, enables the vehicle to dynamically adjust its high-level task plans to avoid unsafe behaviors. 
}\label{fig:plans}
\end{figure}

\vspace{.8em}
\noindent \textbf{Illustrative Example:}
Fig.~\ref{fig:plans} presents an illustrative example. 
TMPUD starts with using our optimal task planner to compute \emph{Plan A}. 
The vehicle takes the first symbolic action from \emph{Plan A} (trajectory in blue color), and executes the action using our motion planner. 
Getting close to \emph{Area 1}, the vehicle plans to \emph{merge left}. 
However, the safety estimator at the motion level reports a low safety value in \emph{Area 1}. 
This computed safety value is incorporated into task planner, where the task planner integrates the safety value into its cost function, and re-computes an optimal plan, \emph{Plan B}. 
Different from \emph{Plan A}, \emph{Plan B} suggests the vehicle to \emph{go straight}, and \emph{merge left} in \emph{Area 2}. 
In this trial, the vehicle was able to follow \emph{Plan B} all the way to the goal. 
TMPUD enabled the vehicle to avoid the risky behavior of merging left in \emph{Area 1} without introducing extra motion cost. A demo video is provided on YouTube.\footnote{\url{https://youtu.be/8NHQYUqMyoI}}

\subsection{Full and Abstract Simulation Platforms}
Experiments conducted in CARLA are referred as being in \textbf{full simulation}. 
All vehicles move at a constant speed ($20 km/h$) on average. 
In full simulation, ego vehicle performs the whole plan at the task level in a the presence of other vehicles. 
We spawn different numbers of vehicles (200 and 120), and refer to traffic of the two environments as being \textbf{heavy} and \textbf{normal} respectively. 

Running full simulation using CARLA is time-consuming, preventing us from conducting large-scale experiments. 
For instance, results reported in this paper are based on tens of thousands of experimental trials, and \emph{full simulation in this scale would have required months of computation time}. 
To conduct large numbers of experimental trials, we developed an \textbf{abstract simulation} platform, where action outcomes are sampled from pre-computed probabilistic world models. 
Parameters of the world models (for abstract simulation) are learned by repeatedly spawning the ego and surrounding vehicles in a small area, and statistically analyzing the results of the vehicles' interaction.

In particular, we spent the most effort in analyzing the outcomes of ``merging lane'' actions due to its significant potential risks. 
We empirically computed the probabilities of the three different outcomes of ``merging lane'' actions, including ``merge'', ``collide'', and ``stop''. 
We introduced \textbf{two domain factors} into the abstract simulation platform, including \textbf{density} and \textbf{acceleration}. 
In high-density environments, the ego vehicle is surrounded by three vehicles, while this number is reduced to one in low-density environments. 
In high-acceleration environments, surrounding vehicles' acceleration (in $m/s^2$) is randomly sampled in $[-1.0, 1.0]$, while this range is $[-0.5, 0.5]$ in low-acceleration environments. 


\begin{figure*}
\centering
\includegraphics[width=0.55\columnwidth]{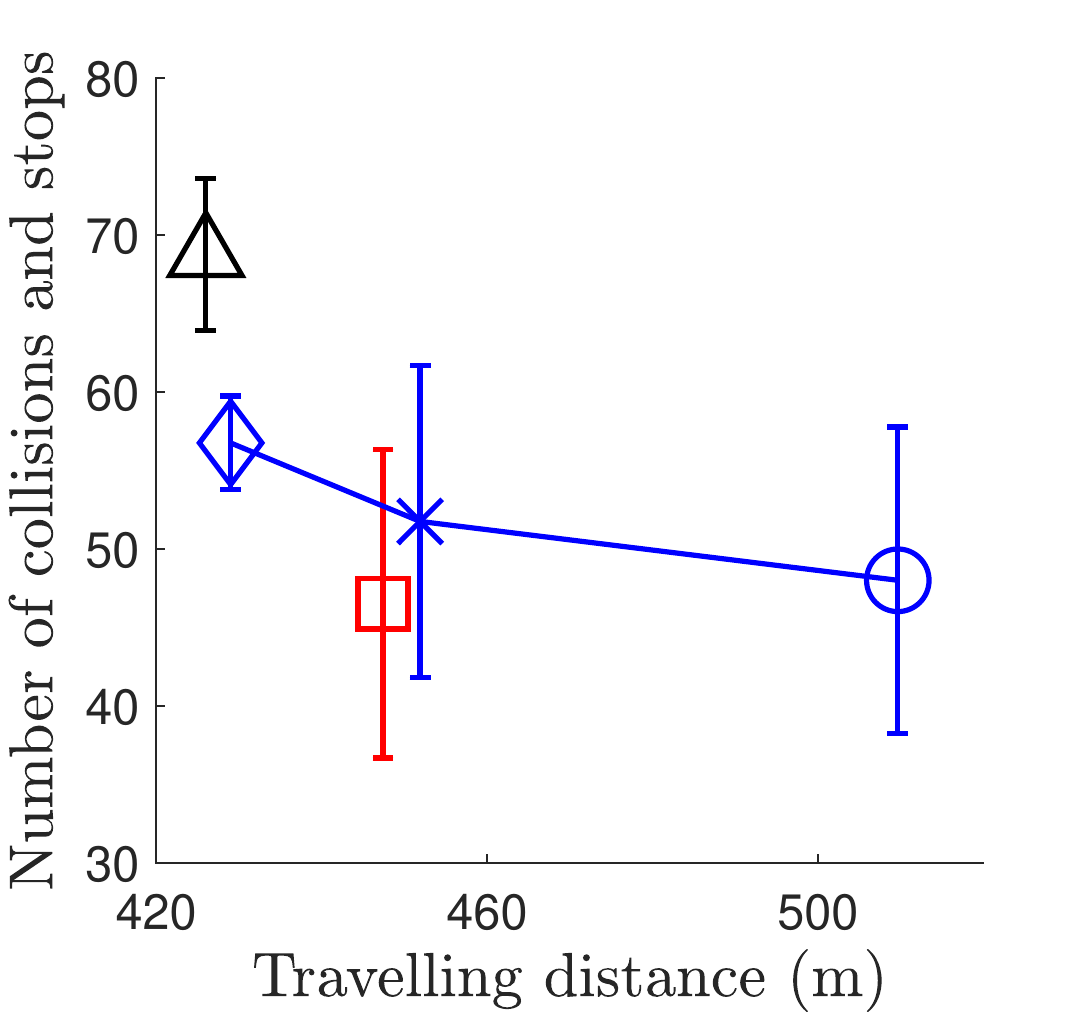}\hspace{-0.5cm}
\includegraphics[width=0.55\columnwidth]{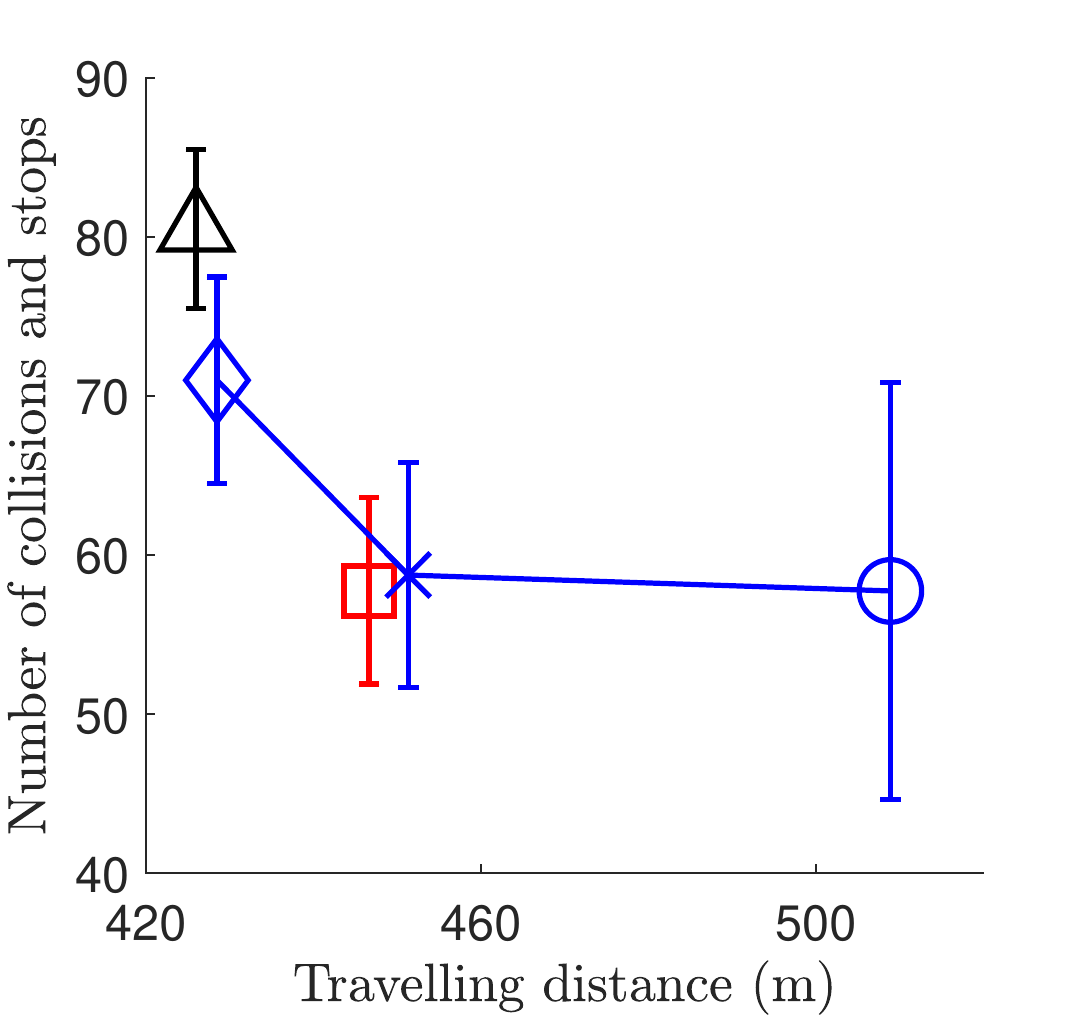}\hspace{-0.5cm}
\includegraphics[width=0.55\columnwidth]{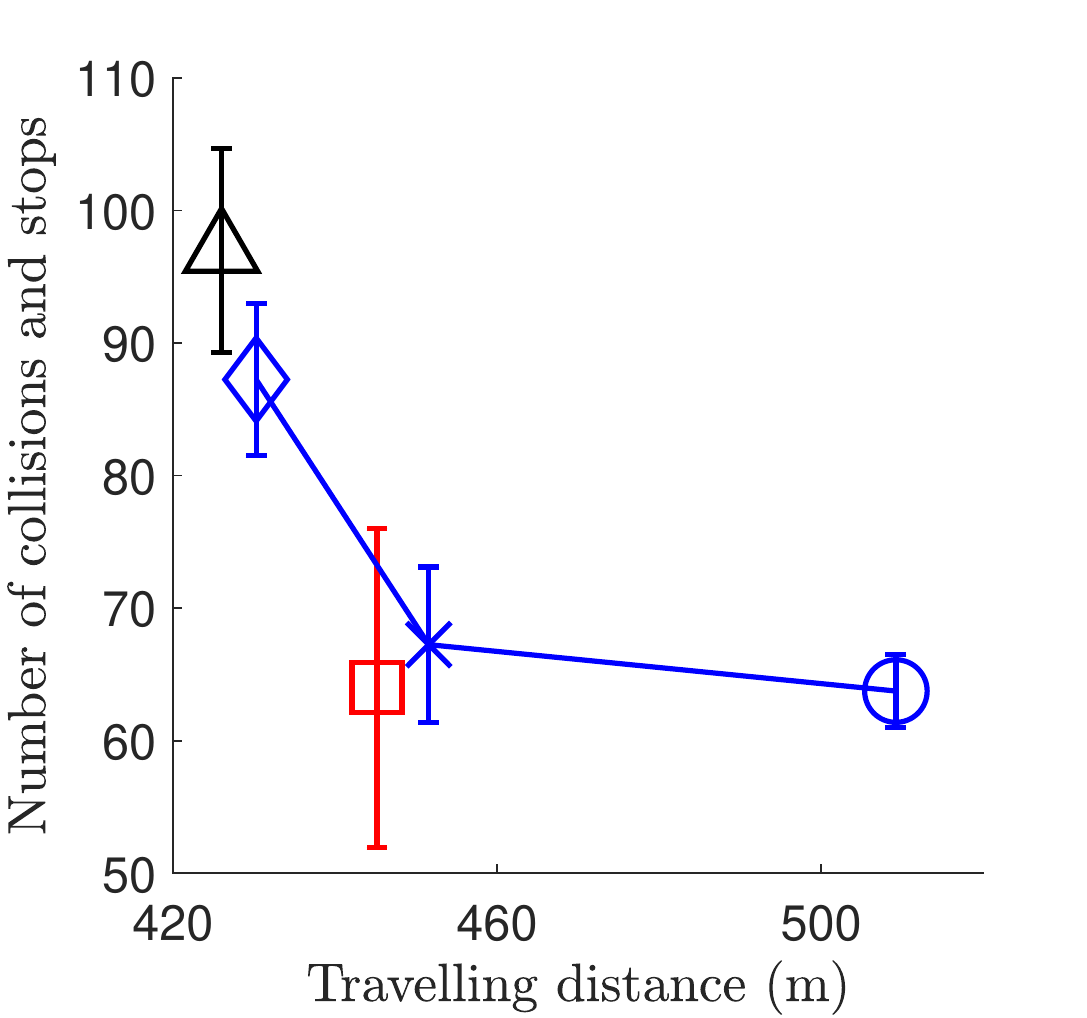}\hspace{-0.5cm}
\includegraphics[width=0.55\columnwidth]{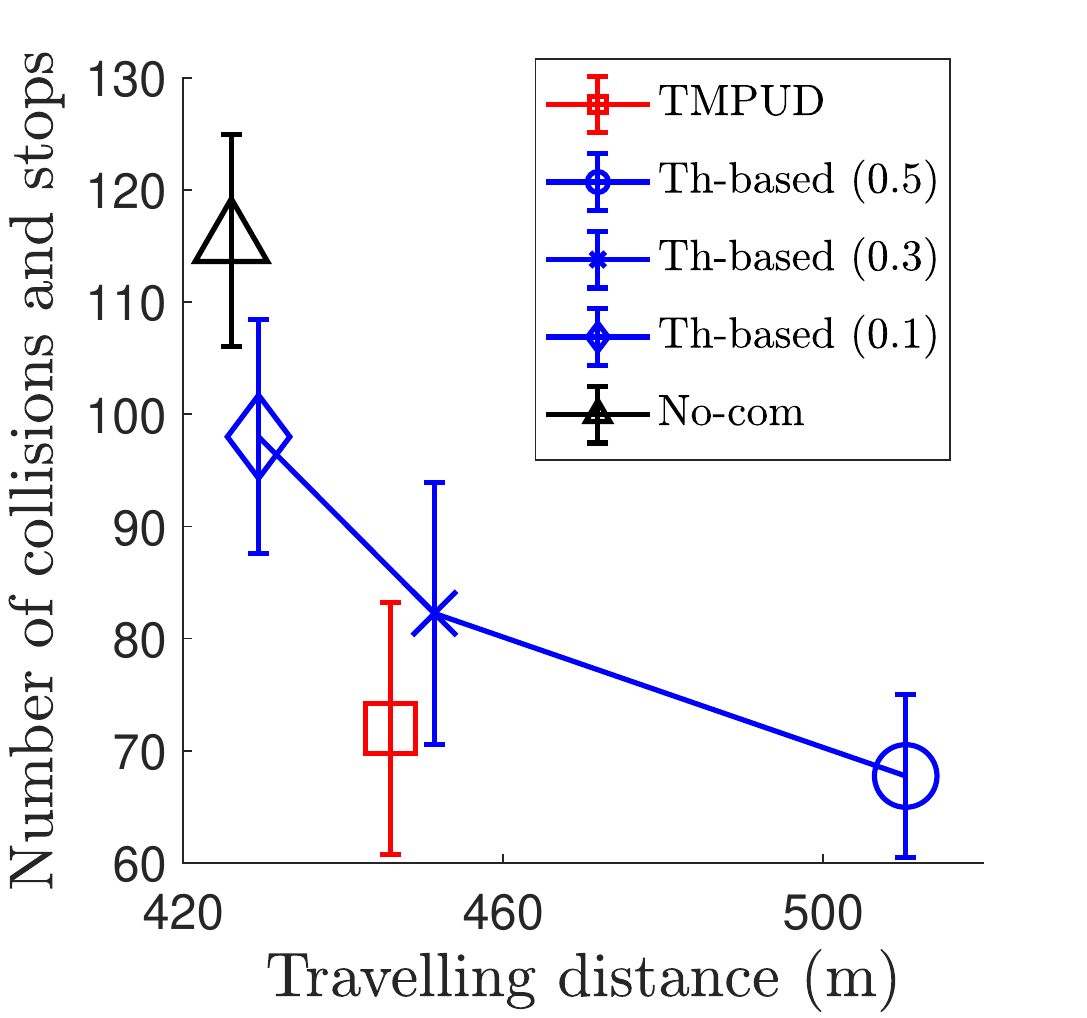}\hspace{-0.5cm}
\caption{\textbf{Abstraction simulation:} the overall performances of TMPUD and two baseline methods.
The x-axis represents the average traveling distance, and the y-axis represents the total number of collisions and stops. 
The four subfigures correspond to four different road conditions. 
The road conditions, from left to right, are \textbf{low-density and low-acceleration}, \textbf{low-density and high-acceleration}, \textbf{high-density and low-acceleration}, \textbf{high-density and high-acceleration}. 
Under each road condition, we evaluate each algorithm using 4000 trials. 
We did batch-based evaluations with four batches for significance analysis, where each batch includes 1000 trials. 
}\label{fig:mainresult}
\end{figure*}

\subsection{Evaluation Metrics and Two Baseline Methods}
The goal of TMPUD is to improve task-completion efficiency (to reduce traveling distance), while guaranteeing safety. 
So, the two most important evaluation metrics are \textbf{traveling distance} and \textbf{the number of unsafe behaviors}, where unsafe behaviors cause either collisions or force at least one surrounding vehicle to stop (to avoid collisions). 

The two baseline methods used in this research are selected from the literature, and referred to as \emph{No-communication} (\textbf{No-com}), and \emph{Threshold-based} (\textbf{Th-based}). 
The No-com baseline~\cite{chen2015task} forces the vehicle to execute all task-level actions at the motion level, while driving behaviors' safety values are not considered. 
The Th-based baseline~\cite{srivastava2014combined} enables the motion planner to ``reject'' a task-level action when its safety value is lower than a threshold $\beta$, where a higher (lower) $\beta$ threshold makes a vehicle more conservative (aggressive). 
In case of an action being rejected, the task planner will compute a new plan to avoid the risky action. 
We develop \textbf{three versions} of of the Th-based baseline with different $\beta$ values (0.1, 0.3, and 0.5).

\begin{table}[t]
\vspace{.8em}
\centering
\caption{\textbf{Full simulation}: Traveling distance and number of collisions and stops for three algorithms under  different traffic conditions (normal and heavy traffic).}\label{tab:full}
\begin{tabular}{|c|c|c|c|c|}
\hline
\multirow{6}{*}{\begin{tabular}[c]{@{}c@{}}Normal\\ Traffic\end{tabular}} & \multicolumn{2}{c|}{Algorithm}                                                  & \begin{tabular}[c]{@{}c@{}}Traveling\\ Distance (m)\end{tabular} & \begin{tabular}[c]{@{}c@{}}Num. of collisions \\ and stops\end{tabular} \\ \cline{2-5} 
                                                                          & \multicolumn{2}{c|}{TMPUD}                                                      & 514                                                             & 0                                                          \\ \cline{2-5} 
                                                                          & \multirow{3}{*}{\begin{tabular}[c]{@{}c@{}}Th-based
                                                                          \end{tabular}} & $\beta = 0.5$ & 537                                                             & 0                                                            \\ \cline{3-5} 
                                                                          &                                                                            & $\beta = 0.3$ & 513                                                            & 5                                                            \\ \cline{3-5} 
                                                                          &                                                                            & $\beta = 0.1$ & 478                                                             & 24                                        \\ \cline{2-5} 
                                                                          & \multicolumn{2}{c|}{\begin{tabular}[c]{@{}c@{}}No-com\end{tabular}} & 426                                                             & 48                                                          \\ \hline
\end{tabular}
\begin{tabular}{|c|c|c|c|c|}
\hline\hline
\multirow{6}{*}{\begin{tabular}[c]{@{}c@{}}Heavy\\ Traffic\end{tabular}} & \multicolumn{2}{c|}{Algorithm}                                                  & \begin{tabular}[c]{@{}c@{}}Traveling\\ Distance (m)\end{tabular} & \begin{tabular}[c]{@{}c@{}}Num. of collisions \\ and stops\end{tabular} \\ \cline{2-5} 
                                                                          & \multicolumn{2}{c|}{TMPUD}                                                      & 530                                                             & 2                                                           \\ \cline{2-5} 
                                                                          & \multirow{3}{*}{\begin{tabular}[c]{@{}c@{}}Th-based\end{tabular}} & $\beta = 0.5$ & 545                                                             & 2                                                            \\ \cline{3-5} 
                                                                          &                                                                            & $\beta = 0.3$ & 528                                                            & 7                                                            \\ \cline{3-5} 
                                                                          &                                                                            & $\beta = 0.1$ & 497                                                             & 35                                        \\ \cline{2-5} 
                                                                          & \multicolumn{2}{c|}{\begin{tabular}[c]{@{}c@{}}No-com\end{tabular}} & 426                                                             & 54                                                          \\ \hline
\end{tabular}
\vspace{-.3em}
\end{table}

\subsection{Results}
\label{sec:results}


\noindent \textbf{Results from Full Simulation: } 
Table~\ref{tab:full} presents the results in comparing TMPUD to the two baseline methods. 
From the table, we see that, in both road conditions, TMPUD achieved the lowest traveling distance, in comparison to those methods that produced compared safety levels (in terms of the number of collisions and stops). 
For instance, under normal traffic, only the Th-based baseline with $\beta=0.5$ was able to completely avoid collisions and stops, but it produced an average traveling distance of $537m$. 
In comparison, TMPUD required only $514m$, while completely avoided collisions and stops. 
Under heavy traffic, TMPUD (again) produced the best performance in safety (based on the number of collisions and stops), while requiring less traveling distance in comparison to the only baseline (\emph{Th-based} with $\beta=0.5$) that produced comparable performance in safety. 
The experimental trials (200 for each approach) from full simulation took eight full workdays. 
We aim at evaluating the performance of TMPUD under different domain factors, requiring a much larger number of trials, which motivated us to conduct experiments using the abstract simulator.


\noindent \textbf{Results from Abstract Simulation:}
Fig.~\ref{fig:mainresult} presents the performances of TMPUD and the baseline methods in both traveling distance and the number of unsafe behaviors. 
The x-axis corresponds to the average traveling distance, and y-axis corresponds to the total number of collisions and stops (both are considered failure cases of driving behaviors). 
From the four subfigures, we see that TMPUD is the most efficient (x-axis) among those methods that produced comparable performances in safety (y-axis), except that \emph{Th-based} ($\beta=0.5$) produced slightly less unsafe behaviors (but it performed poorly in efficiency). 

There are a few side observation. 
Not surprisingly, \emph{No-com} produced the worst performance of in safety (y-axis), though its traveling distance remains the lowest. 
This is because, using \emph{No-com}, the vehicle blindly executes task-level actions while unrealistically believing driving behaviors are always safe. 
The \emph{Th-based} baseline's performance depends on its safety threshold ($\beta$), where a greater value produces safer but less efficient behaviors. 
The results support our claim that TMPUD improves vehicles' task-completion efficiency, while ensuring safety in different road conditions. 

\section{Conclusions and Future work}
In this paper, focusing on urban driving scenarios, we develop a safety evaluation algorithm, and a task-motion planning algorithm, called TMPUD, for autonomous driving. 
TMPUD, for the first time, bridges the gap between task planning and motion planning in autonomous driving.
We have extensively evaluated TMPUD using a 3D urban driving simulator (CARLA) and an abstract simulator. 
Results suggest that TMPUD improves the task-completion efficiency in different road conditions, while ensuring the safety of driving behaviors. 

In the future, we will implement TMPUD using different task and motion planners, and evaluate their performances in different testing platforms (e.g., using simulators with a physics engine) under different conditions. 
Also, there is the possibility of implementing and evaluating TMPUD using indoor mobile robots. 


\section*{Acknowledgements}
\noindent
The authors thank Fangkai Yang for discussions on this topic. This work has taken place in the Autonomous Intelligent Robotics (AIR) Group at SUNY Binghamton. AIR research is supported in part by grants from the National Science Foundation (IIS-1925044 and REU Supplement), Ford Motor Company (URP Awards), OPPO (Faculty Research Award), and SUNY Research Foundation.

\bibliography{ref}{}
\bibliographystyle{IEEEtran}

\end{document}